\documentclass{article}

\usepackage{arxiv}

\usepackage[utf8]{inputenc} 
\usepackage[T1]{fontenc}    
\usepackage{hyperref}       
\usepackage{url}            
\usepackage{booktabs}       
\usepackage{amsfonts}       
\usepackage{nicefrac}       
\usepackage{microtype}      
\usepackage{lipsum}
\usepackage{indentfirst}
\usepackage{graphicx}
\usepackage{multicol, multirow}
\usepackage{caption} 
\usepackage{subcaption}
\usepackage{float}
\usepackage{aliascnt}
\newaliascnt{eqfloat}{equation}
\newfloat{eqfloat}{h}{eqflts}
\floatname{eqfloat}{Equation}

\bgroup

\newcommand*{\ORGeqfloat}{}
\let\ORGeqfloat\eqfloat
\def\eqfloat{%
  \let\ORIGINALcaption\caption
  \def\caption{%
    \addtocounter{equation}{-1}%
    \ORIGINALcaption
  }%
  \ORGeqfloat
}

\title{A novel approach to remove foreign objects from chest X-ray images}

\author{
 Hieu X. Le $^{1,\dagger}$, Phuong D. Nguyen $^{1,\dagger}$, Thang H. Nguyen $^1$, Khanh N.Q. Le $^{2,3,*}$, Thanh T. Nguyen $^{1,*}$ 
  \\
  $^1$ AI Research Laboratory, RE-THINKING COMPANY, Vietnam
  \\
  $^2$ Professional Master Program in Artificial Intelligence in Medicine,\\ College of Medicine, Taipei Medical University, Taipei City 106, Taiwan
  \\
  $^3$ Research Center for Artificial Intelligence in Medicine,\\Taipei Medical University, Taipei City 106, Taiwan
  \\
  $^\dagger$ Hieu and Phuong contributed equally to this work.
  \\
  $^*$ Corresponding authors: khanhlee@tmu.edu.tw, thanh.steve@rethinkingai.com
}
\date{}

\begin{document}
\maketitle

\begin{abstract} We initially proposed a deep learning approach for foreign objects inpainting in smartphone-camera captured chest radiographs utilizing the cheXphoto dataset. Foreign objects which can significantly affect the quality of a computer–aided diagnostic prediction are captured under various settings. In this paper, we used multi-method to tackle both removal and inpainting chest radiographs. Firstly, an object detection model is trained to separate the foreign objects from the given image. Subsequently, the binary mask of each object is extracted utilizing a segmentation model. Each pair of the binary mask and the extracted object are then used for inpainting purposes. Finally, the in-painted regions are now merged back to the original image, resulting in a clean and non-foreign-object-existing output. To conclude, we achieved state-of-the-art accuracy. The experimental results showed a new approach to the possible applications of this method for chest X-ray images detection.

\end{abstract}


\section{Introduction}
\begin{figure}[htbp]
\begin{subfigure}{.5\textwidth}
  \centering
  \includegraphics[width=.6\linewidth, height=0.7\linewidth]{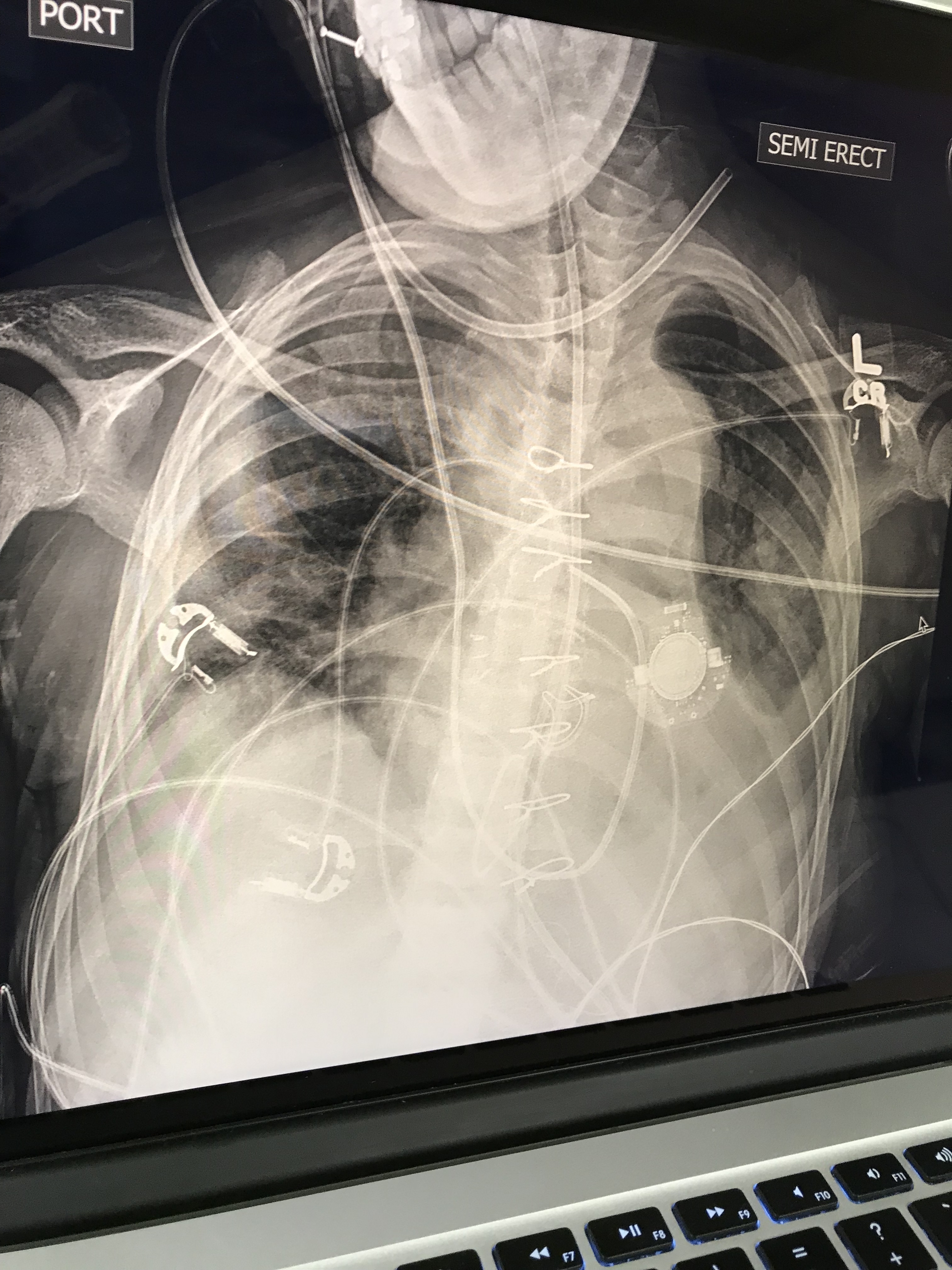}  
\end{subfigure}
\begin{subfigure}{.5\textwidth}
  \centering
  \includegraphics[width=.6\linewidth, height=0.7\linewidth]{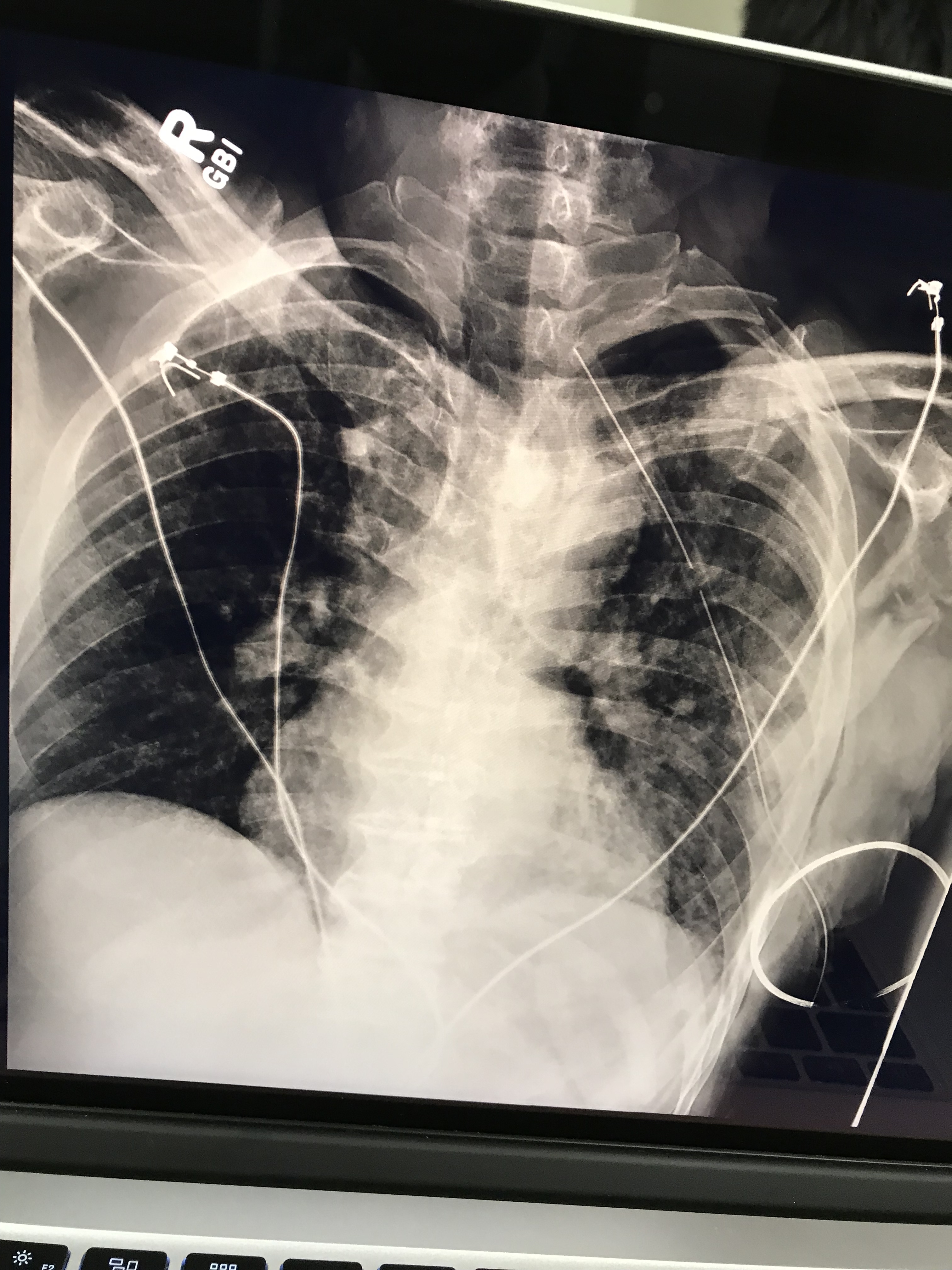}
\end{subfigure}
\caption{CheXphoto images were captured from smartphone}
\label{fig:chexphoto-dataset}
\end{figure}

\parindent=2em
The performance of the computer-aid diagnosis approach plays an important role in clinical care treatment \cite{cahan2017learning}. In which, chest x-ray is considered the most innovative technology for health services because of the possibility of application of several breakthrough technologies in artificial intelligence \cite{jaeger2013automatic}. AI can do this task quickly, and cheaply that can significantly improve diagnostics and ultimately treat the disease through chest radiographs. However, in many situations, poor quality images with complicated foreign objects are usually directly unexpected results. In which, foreign objects are a major hindrance to remove these factors which affect the performance of interpretation of chest radiographs. For instance, foreign elements are shown in figure \ref{fig:chexphoto-dataset}.

Xue \cite{xue2015foreign} has the first novel technology for identification common foreign objects in chest radiographs that are linked four steps including normalization, image enhancement, segmentation, and button object. They performed circular hough transform (CHT) and Viola Jones algorithm. For CHT, this model has achieved 83\% precision and 85\% recall, by contrast of Viola Jones algorithm with 86\% precision and 94\% recall. Similarly, Zohora \cite{zohora2017foreign} has also applied multi-step including normalization and adjustment for pre-processing chest x-ray images. Next, they used edge detection to circle segmentation candidates and performed CHT to detect foreign object element. The results have achieved significantly including 96\% precision, 90\% recall, and 90\% F1 score. In 2018, they have shown a new approach to collect foreign elements-like unsupervised clustering with more than 90\% accuracy \cite{zohora2018circle}. Another method used a k-NN classifier algorithm was applied by Hogeweg \cite{hogeweg2013foreign} to remove different types of foreign objects such as brassier clips, jewelry, pacemakers, etc.

However, these current models are no longer carried out in x-ray photography capturing cell phones which is one of the most challenging due to under various settings. In addition, only button objects were implemented in previous studies. In this paper, we deployed on CheXphoto datasets \cite{phillips2020chexphoto} that  were created with multi-complicated foreign objects. In which, 10,507 photo x-ray images of 3,000 unique patients for training set that get a better insight into developing the detection model with high accuracy on smartphones. 

The first contribution of our proposed study is the novelty of the method. To the best of our knowledge, this is the first time a multi-step Deep Learning-based study has been carried out in order to handle foreign objects in chest radiographs captured by smartphone cameras. In addition, we achieved an acceptable accuracy of 90\%, while other metrics including IoU, Dice scores reached up to 81 and 90\% respectively. Another contribution of this study is that our study proposed a list of annotation criteria as long as the cross-validation label reviewing process to allow the consistency in annotating and reviewing the label. Finally, this study can serve as assistance in improving the performance of models in lung disease classification problems.

The paper is structured as follows: Firstly, we briefly introduced previous studies for handling foreign objects and then we proposed methodology in five steps including data preparation, pre-processing, foreign object detection, segmentation and inpainting. Secondly, the experimental result section provided the achievement of our pipeline.  Consequently, the conclusion was drawn in the discussion and future work section. 

\section{Experiment Design}
\begin{figure}[htbp]
    \centering
    \includegraphics[width=\linewidth]{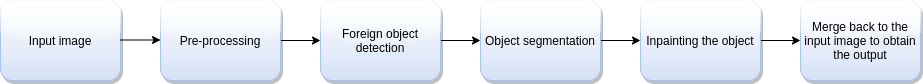}
    \caption{The proposed pipeline for our solution}
    \label{fig:pipeline}
\end{figure}

\begin{figure}[h]
    \centering
    \includegraphics[width=\linewidth]{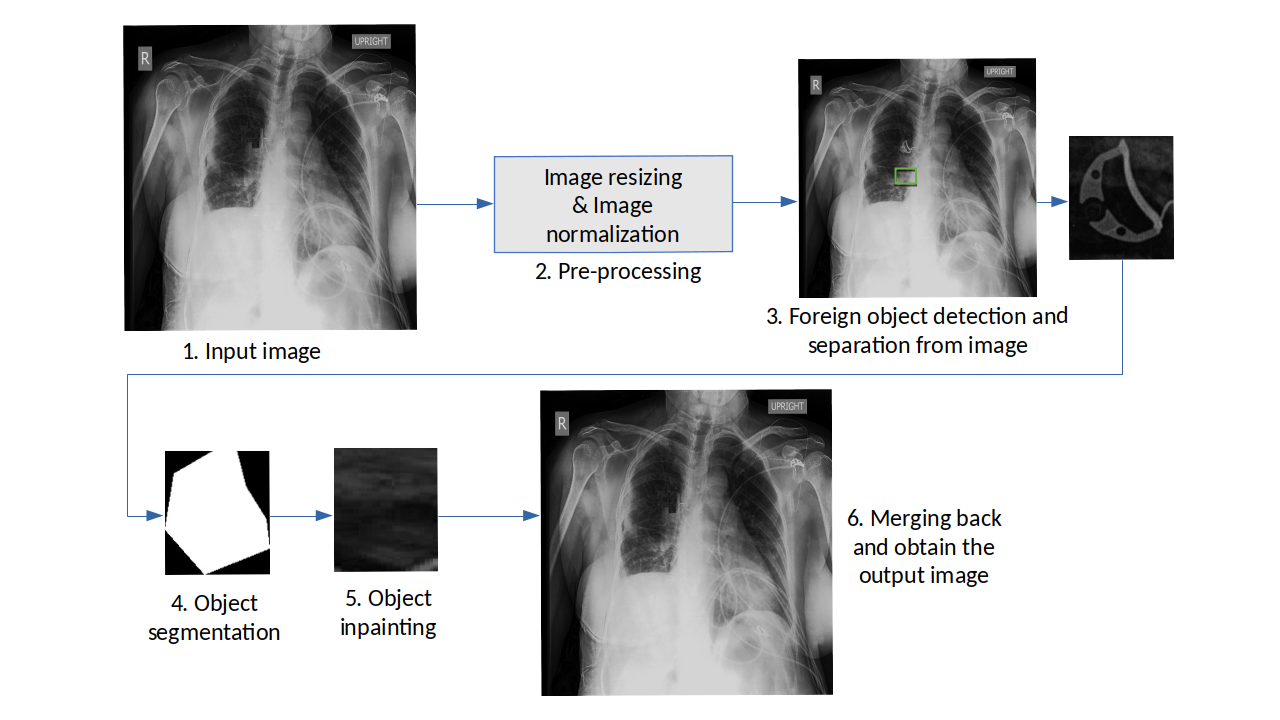}
    \caption{The actual processes given a raw input image.}
    \label{fig:actual-pipeline}
\end{figure}

In this study, we proposed a novel approach to remove foreign objects from initial chest radiographs guided by object detection and segmentation results. Figure \ref{fig:pipeline} and \ref{fig:actual-pipeline} describes 4 main stages included in the approach. In the following section, we will describe each step in detail including the pre-processing techniques, foreign object detection, detail segmentation and inpainting method.

\subsection{Data preparation}
Our experiment was carried out on  a subset of the cheXphoto dataset, including four types of lung diseases: pneumothorax, effusion, pneumonia, atelectasis without uncertain labels for all (Table \ref{tab:data-stat}). These chest x-ray images were collected from the cheXphoto dataset with natural and synthetic transformation, removed lateral style. Afterward, we continually filtered out non-foreign object images. The remaining images with complicated foreign objects were applied to the training model.

To prove the contribution of the inpainting technique in improving the performance of classification models, we adopted the second experiment and constructed  another dataset (Table \ref{tab:test-train}). Samples from this dataset were selected from the original chexPhoto dataset. To enable independence from the inpainting dataset and avoid any possible bias or imbalance, samples were selected in the way such that, there is no overlap between the inpainting dataset and the classification dataset, moreover, the ratio between both classes in the dataset remains balanced.

\begin{table}[htbp]
    \centering
    \begin{tabular}{|l|r|r|r|r|r|}
    \hline
                   & \multicolumn{1}{l|}{Pneumonia} & \multicolumn{1}{l|}{Atelectasis} & \multicolumn{1}{l|}{Pneumothorax} & \multicolumn{1}{l|}{Pleural Effusion} & \multicolumn{1}{l|}{No finding} \\ \hline
    Train set & 32 & 359 & 251 & 950 & 132\\ \hline
    Validation set & 14 & 94 & 49 & 202 & 33 \\ \hline
    Test set & 8 & 99 & 74 & 189 & 38 \\ \hline
    \end{tabular}
    \caption{Number of chest x-ray images for training, test and validation}
    \label{tab:data-stat}
\end{table}

\begin{table}[htbp]
    \centering
    \begin{tabular}{|l|r|r|r|}
    \hline
        & Normal & Abnormal \\ \hline
    Train set & 196 & 200\\ \hline
    Test set & 66 & 56\\ \hline
    \end{tabular}
    \caption{Number of chest x-ray images for training, test set used for the classification task}
    \label{tab:test-train}
\end{table}

\textbf{Annotations method}

By default, the CheXphoto dataset does not include any annotations for object detection or segmentation. Thus, we have to determine the standards of the annotation criteria and the reviewing process for both object detection and segmentation tasks. To ensure the consistency and accuracy in annotations, we thereby follow a cross-validation reviewing progress, where each image has to be reviewed by at least two people. Each image is scanned to identify if it follows all of the below criteria mentioned in table \ref{tab:standard-annotation}. If any criteria are disobeyed, the image’s annotations are cleared completely, and the image is put back to the queue of annotation and reviewed again.

\begin{table}[htbp]
    \begin{tabular}{|l|l|}
    \hline
    Criteria & Requirements \\ \hline
    Shape of objects 
    & 
    \vtop{\hbox{\strut Sickle or round or earbud-like shaped objects}\hbox{\strut 
    \begin{tabular}{p{2cm}p{2cm}p{2cm}}
        \begin{minipage}{0.33\textwidth}
          \includegraphics[width=0.25\linewidth, height=0.3\linewidth]{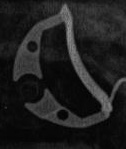}
        \end{minipage}
        &
        \begin{minipage}{0.33\textwidth}
          \includegraphics[width=0.25\linewidth, height=0.3\linewidth]{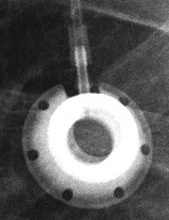}
        \end{minipage}
        &
        \begin{minipage}{0.33\textwidth}
          \includegraphics[width=0.25\linewidth, height=0.3\linewidth]{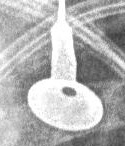}
        \end{minipage}
    \end{tabular}
    
    }}
    
    \\ \hline
    Location                             & Within and located in the centre of the bounding boxes                             \\ \hline
    Region                               & An object is annotated as long as the object is in or is partly in the lung region \\ \hline
    Number of vertices in a bounding box & At least 6-8 vertices are required for each bounding box                           \\ \hline
    Number of objects                    & The number of objects annotated by each reviewers have to be the same              \\ \hline
    \end{tabular}
    \caption{Standards of the annotation criterias}
    \label{tab:standard-annotation}
\end{table}


\subsection{Pre-processing}
Initially, the chest x-ray images are extracted from the original images. To ensure the consistency in pixel values among different images, we thereby normalize all images to the range of [0-255] in pixel values. The normalized samples are then resized to a predefined size of WxH.

\subsection{Foreign object detection}

In order to separate the regions of interest (ROIs), for instance, the undesired and foreign object, from the pre-processed images, we trained a YOLOv4-based model \cite{bochkovskiy2020yolov4} to detect the rectangles surrounding the targets. Thus, this step enables us to cut out the focused region only, instead of having to work on the whole global and large images. The foreign object detection model is constructed based on the widely-known Yolov4 architecture, which has an initial learning rate of 0.00261, batch size of 4 over 100 epochs. The loss function used is the GIoU loss. After this step, every single foreign object is separated into an image, which is then fed to the segmentation model.

\subsection{Foreign object segmentation}

For the segmentation task, we adopt a UNET-like structure \cite{ronneberger2015u} for the segmentation model. Instead of using the vanilla encoder, we made use of the popular EfficientNet-B0 model as the chosen encoder \cite{tan2019efficientnet}. We used the custom BCE-Dice loss as the cost function and Dice score and Intersection over Union ratio as learning metrics. The model is trained for 100 epochs, with a learning rate of 0.01 and a batch size of 8. Upon the completion of this step, the binary masks for foreign objects are generated, in which positive pixels are marked equal to the maximum intensity values (255) otherwise, negative pixels are marked as minimum intensity values (0).

\subsection{Foreign object inpainting}

The inpainting technique enables the removal of unwanted objects and the reconstruction of the hidden part in the target images. In this study, we used the Fast Marching algorithm to tackle the challenges of having noisy, undesired foreign objects in the chest x-rays. The algorithm inpainted a pixel by assigning it with a normalized weighted sum of all its surrounding neighbor pixels. We determined the size of the circular neighborhood surrounding every considered pixel by the radius size of T. After this step, the foreign objects should be completely cleared from the ROI images. It is worth mentioning that the in-painted ROIs are then merged back to the original images.

\subsection{Evaluation metrics}

To evaluate the overall performance and the efficiency of the approach, we proposed several metrics for deeper analysis. We proposed the mean difference in object counting (MDOC) (Eq. \ref{eq:MDOC}) to quantify the object detection ability of the model. Subsequently, the average Dice score (avgDice) (Eq. \ref{eq:avgDice}) between predicted objects and ground truth objects is also employed to assess how overlapped, similar or accurate the predicted objects are, in comparison to the ground-truths, in pixel-level.  Conventional metrics such as accuracy, precision, recall and F1-score are also accounted for better assessment.

\begin{eqfloat}
    \begin{equation}
        MDOC = \frac{1}{\sum_i^N C_i} \times \sum_i^N (C_i - \hat{C}_i)
    \end{equation}
    \caption{Formula of the mean difference in object counting metric}
    \label{eq:MDOC}
\end{eqfloat}

$C_i$ denotes the total number of foreign objects existing in sample $i^{th}$, $\hat{C}_i$ denotes the total number of predicted foreign objects existing in sample $i^{th}$

\begin{eqfloat}
    \begin{equation}
        avgDice = \sum_i^N\sum_j^M Dice(B_{ij}, \hat{B}_{ij})
    \end{equation}
    \caption{Formula of the average Dice score metric}
    \label{eq:avgDice}
\end{eqfloat}

Where N, M denotes the number of samples and objects available in the dataset, and each sample respectively. $B_j$ and $B_j$ represents the ground-truth and the predicted binary masks of the object $j^{th}$ in sample $i^{th}$

\section{Result}

\subsection{Object detection}
The results obtained with state-of-the-art object detectors. Our model detector got the below results with the image size set 512x512 and confidence threshold 30\%, IOU threshold 60, batch size when evolution using 16 on the test-set we extracted  from ChexPhoto dataset. Our model has defined that the object foreign on the boundary of lung region is true positive and by contrast that of false positive. The percentage of precision, recall, and F1 was 76\%, 70\%, and 73\%, respectively. Meanwhile, around 61\% for mAP@0.5. Table 3 showed several examples of output metrics in object detection.

\begin{table}[htbp]
    \centering
    \begin{tabular}{|c|c|c|c|c|}
    \hline
    Ground Truth & Precision & Recall & F1-score & mAP@0.5 \\ \hline
    1524         & 76        & 70     & 73       & 61      \\ \hline
    \end{tabular}
    \caption{Result of foreign object detection}
    \label{tab:OD-result}
\end{table}

\begin{figure}[ht]
\begin{subfigure}{.5\textwidth}
  \centering
  \includegraphics[width=.8\linewidth, height=0.7\linewidth]{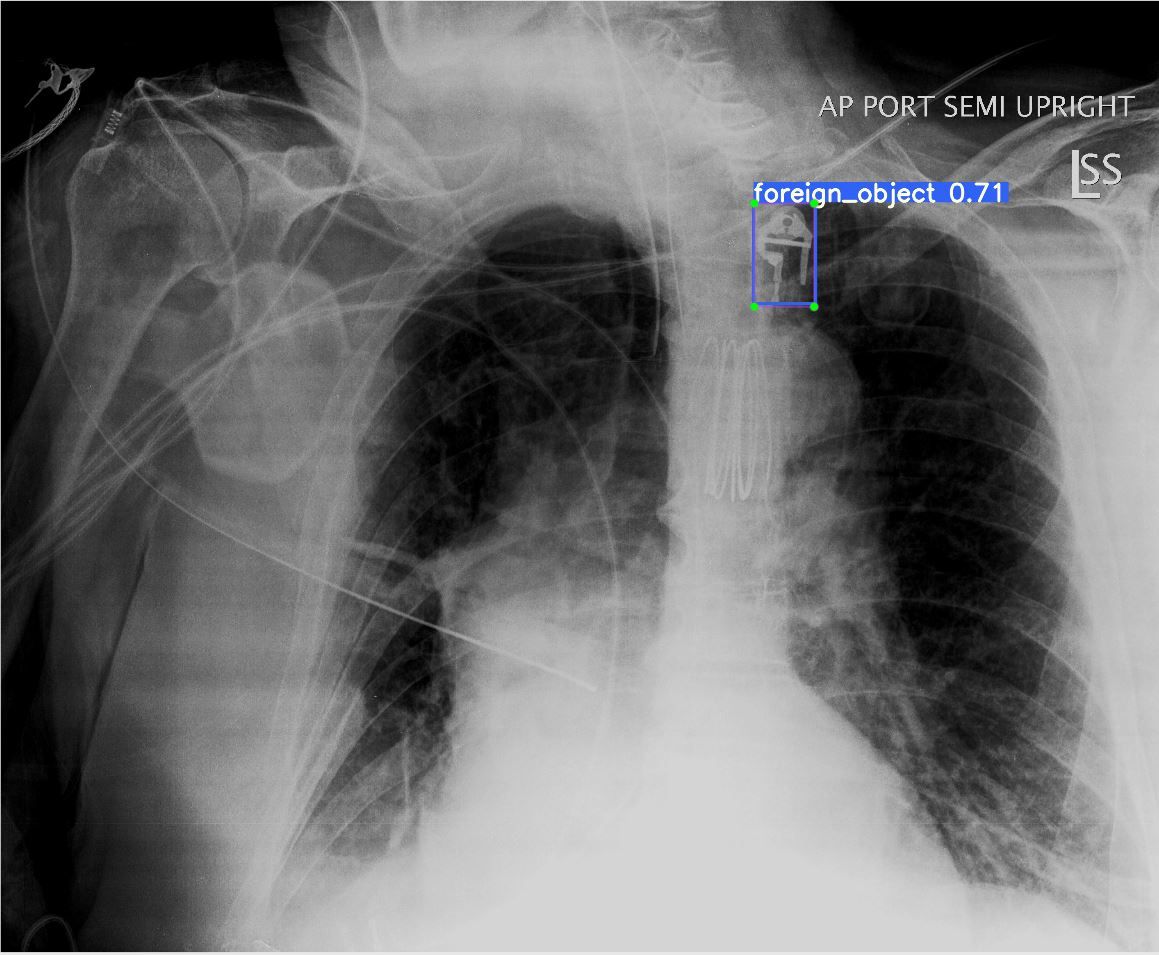}  
\end{subfigure}
\begin{subfigure}{.5\textwidth}
  \centering
  \includegraphics[width=.8\linewidth, height=0.7\linewidth]{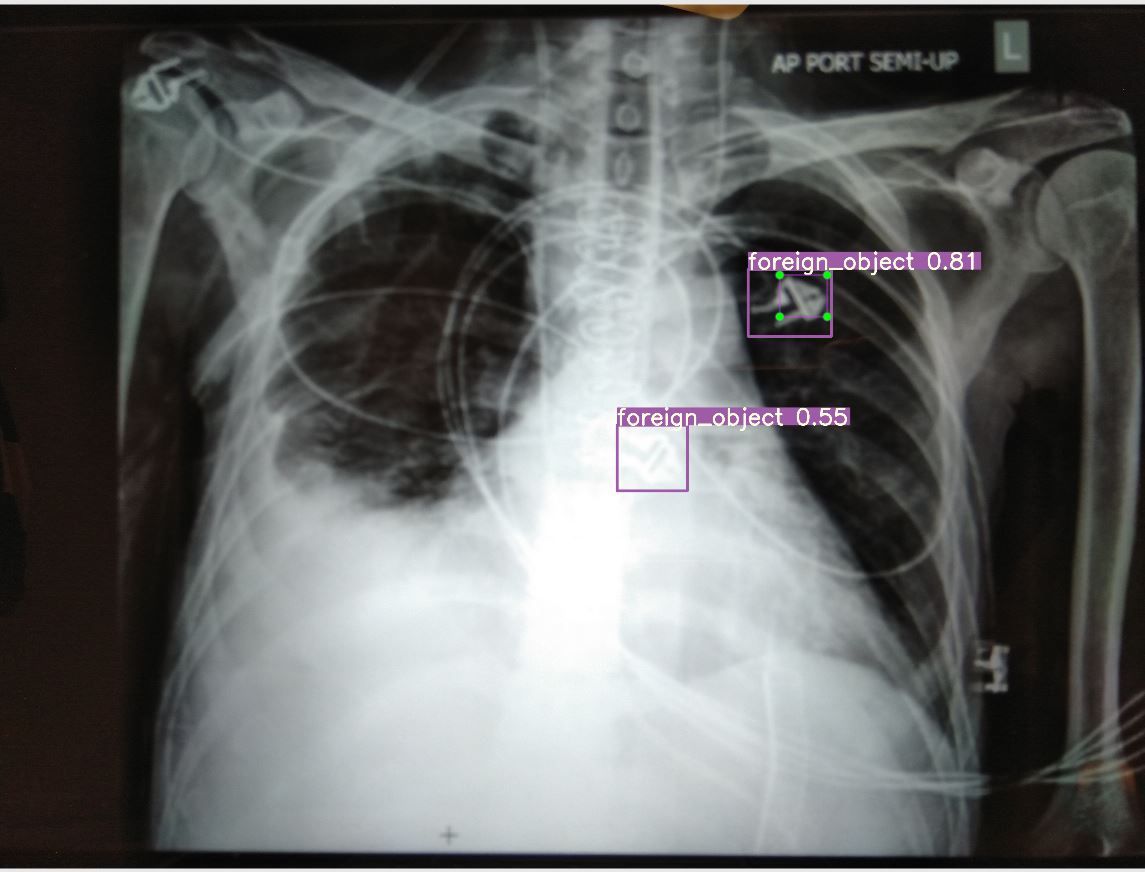}
\end{subfigure}
\caption{True positives of foreigin object detector}
\label{fig:true-positive}
\end{figure}

\begin{figure}[htbp]
\begin{subfigure}{.5\textwidth}
  \centering
  \includegraphics[width=.8\linewidth, height=0.7\linewidth]{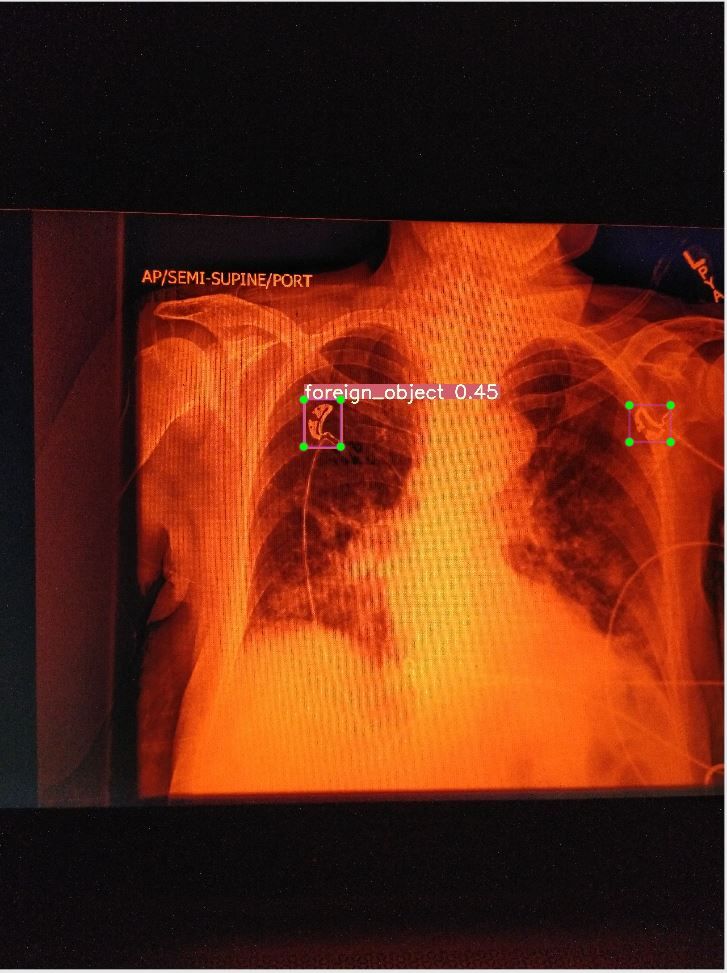}  
\end{subfigure}
\begin{subfigure}{.5\textwidth}
  \centering
  \includegraphics[width=.8\linewidth, height=0.7\linewidth]{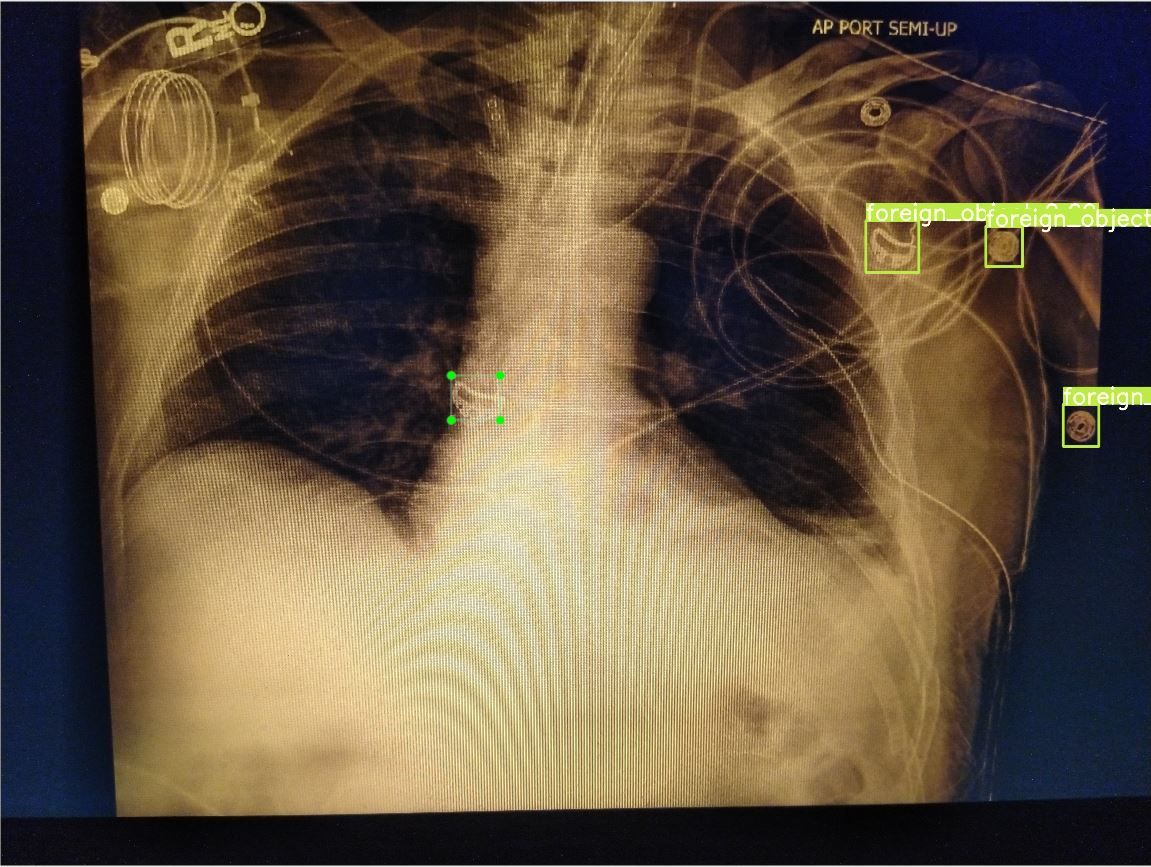}
\end{subfigure}
\caption{False positive of foreign object detector}
\label{fig:false-positive}
\end{figure}

\subsection{Segmentation}
\begin{figure}
    \centering
    \includegraphics[width=0.5\linewidth]{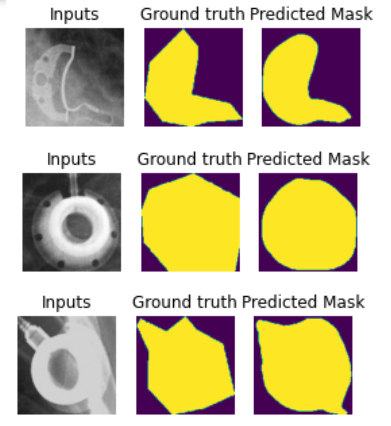}
    \caption{Examples of raw-inputs, ground-truths and binary mask generated from segmentation models}
    \label{fig:result_no_score}
\end{figure}

Table \ref{tab:segment-result} reveals the obtained result utilizing the aforementioned segmentation strategy. The threshold to distinguish between positive pixels and negative pixels is 0.5 by default. Figure \ref{fig:result_no_score} illustrates several examples obtained, where raw inputs and predictions are plotted against the annotated ground-truths.
As a result in table 4, predicted bounding boxes and obtained the level of overlap were considered through Dice and  IoU score@0.5 that have achieved high accuracy with 0.8889, and 0.8045, respectively. The output is closely matched with foreign objects in figure 4. Moreover, pixel-level accuracy@0.5 had a similar tendency of around 93.12\%.

\begin{table}[htbp]
    \centering
    \begin{tabular}{|c|c|c|}
    \hline
    Dice score @0.5 & IoU score @0.5 & Pixel-level accuracy @0.5 \\ \hline
    0.8889 & 0.8045 & 0.9312
    \\ \hline
    \end{tabular}
    \caption{Results of the segmentation task}
    \label{tab:segment-result}
\end{table}

\subsection{Inpainting results}

We examined the performance of the proposed approach by determining the number and the percentage of the objects removed. After processing, the images were reviewed again to check if all foreign objects were inpainted and cleaned completely. If yes, we increased the count, otherwise we continue with the next images. Thus, in the end, we reported in table 6 the number of completely foreign object- inpainted images as long as their corresponding percentage.

To the best of our knowledge, currently, there is an acknowledged lack of quantitative metrics for image inpainting quality evaluation \cite{Dang2013visual}. Conventional metrics are often subjective or only specifically adapted to a particular issue. Thus, we thereby evaluated the usefulness and performance of the proposed approach by assessing the contribution of chest radiograph inpainting to the performance of CNN-based lung abnormality detection models. For every model architecture, we trained the model with 2 versions of the same dataset: the original version and the inpainted version. Other settings and hyperparameters were left unchanged. The models are built based on conventional and state of the art architectures, such as VGG16, Resnet50 and EfficientNet-B2. Performance on the test set of every architecture is then recorded in table 7 for comparison and illustrate the effectiveness of the proposed method in improving the models learning ability.

\begin{table}[htbp]
    \centering
    \begin{tabular}{|c|c|c|c|}
    \hline
    Measurements & Total images available & Number of completely inpainted images & \% of inpainted images \\ \hline
    Values & 501 & 428 & 85 \\ \hline
    \end{tabular}
    \caption{Evaluation of the performance of the object inpainting approach}
    \label{tab:evaluation-result}
\end{table}

\begin{table}[htbp]
\centering

\begin{tabular}{|l|l|l|l|l|l|l|l|l|} 
\hline
\multirow{2}{*}{\begin{tabular}[c]{@{}l@{}}\\Architectures\end{tabular}} & \multicolumn{4}{l|}{Original Dataset} & \multicolumn{4}{l|}{Image Inpainting Dataset}  \\ 
\cline{2-9}
                                                                         & Accuracy & Precision & Recall & F1    & Accuracy & Precsion & Recall & F1              \\ 
\hline
VGG16                                                                    & 0.432    & 0.432     & 0.432  & 0.432 & 0.465    & 0.480    & 0.480  & 0.465           \\ 
\hline
Resnet50                                                                 & 0.642    & 0.681     & 0.662  & 0.633 & 0.680    & 0.690    & 0.690  & 0.680           \\ 
\hline
Efficientnet-b2                                                          & 0.561    & 0.580     & 0.575  & 0.559 & 0.570    & 0.580    & 0.580  & 0.580           \\
\hline
\end{tabular}
\caption{Comparison of original and inpainted data in terms of model performance}
\end{table}

\begin{figure}[htbp]
\begin{subfigure}{.5\textwidth}
  \centering
  \includegraphics[width=.8\linewidth, height=0.7\linewidth]{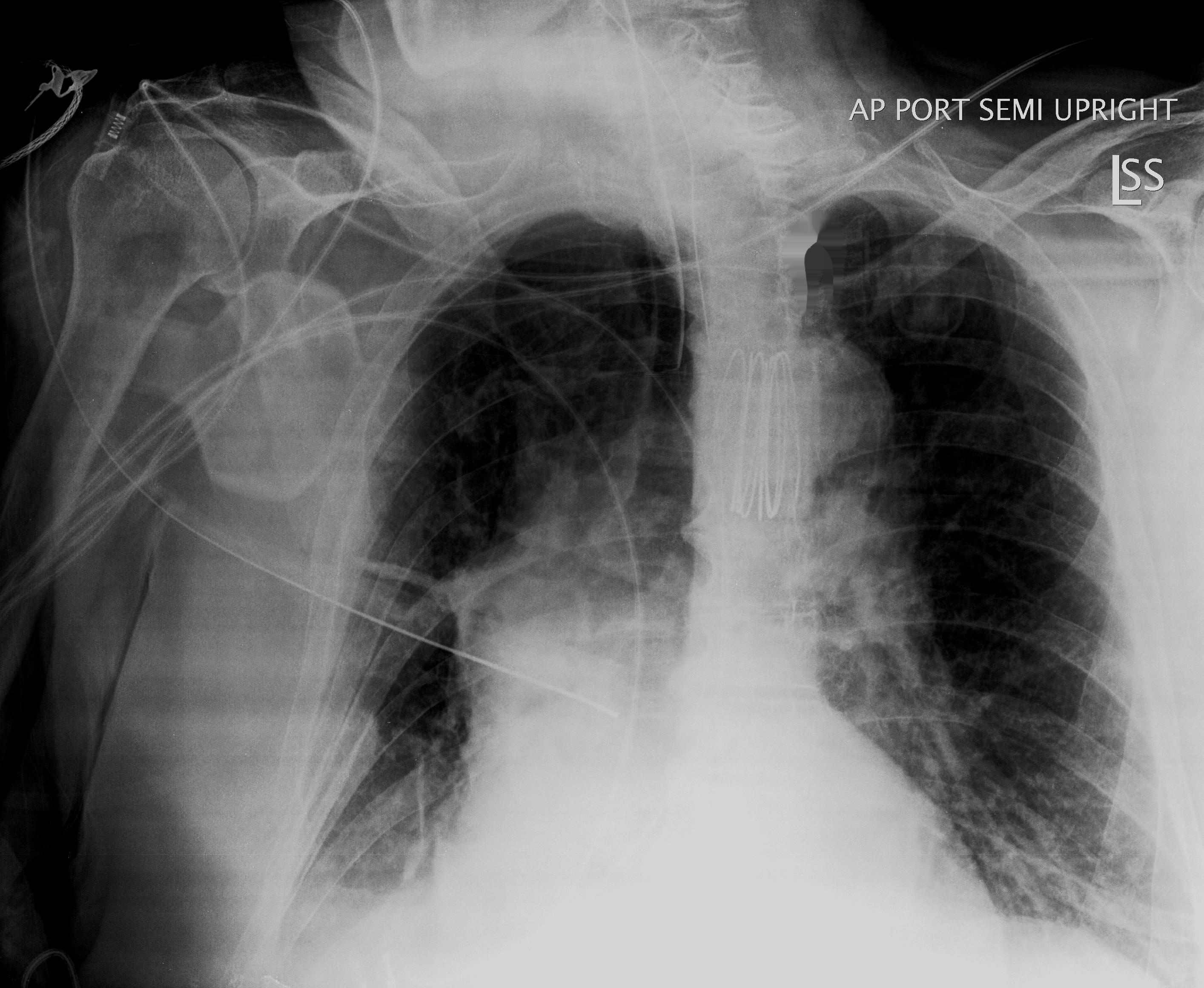}  
\end{subfigure}
\begin{subfigure}{.5\textwidth}
  \centering
  \includegraphics[width=.8\linewidth, height=0.7\linewidth]{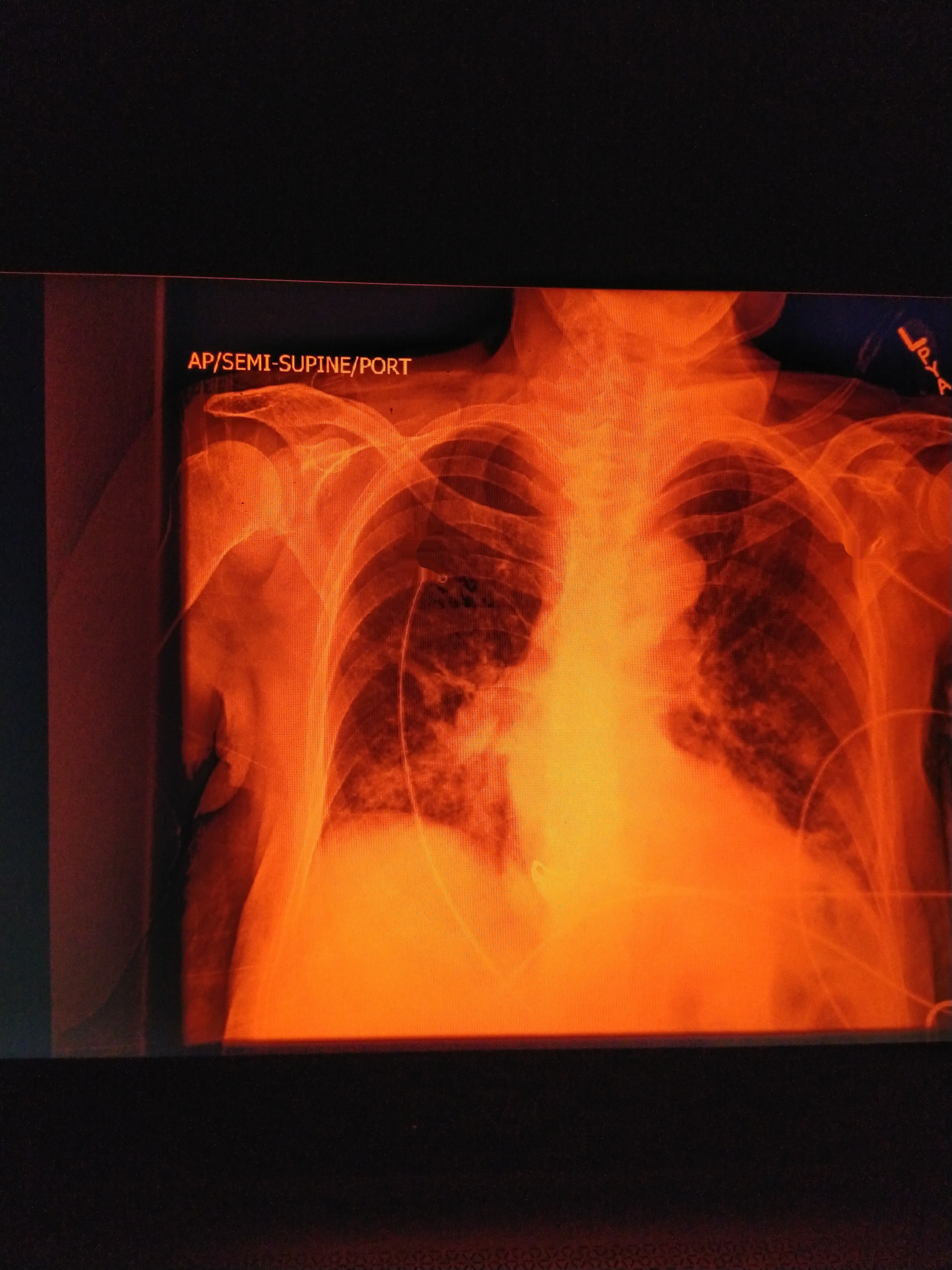}
\end{subfigure}
\begin{subfigure}{.5\textwidth}
  \centering
  \includegraphics[width=.8\linewidth, height=0.7\linewidth]{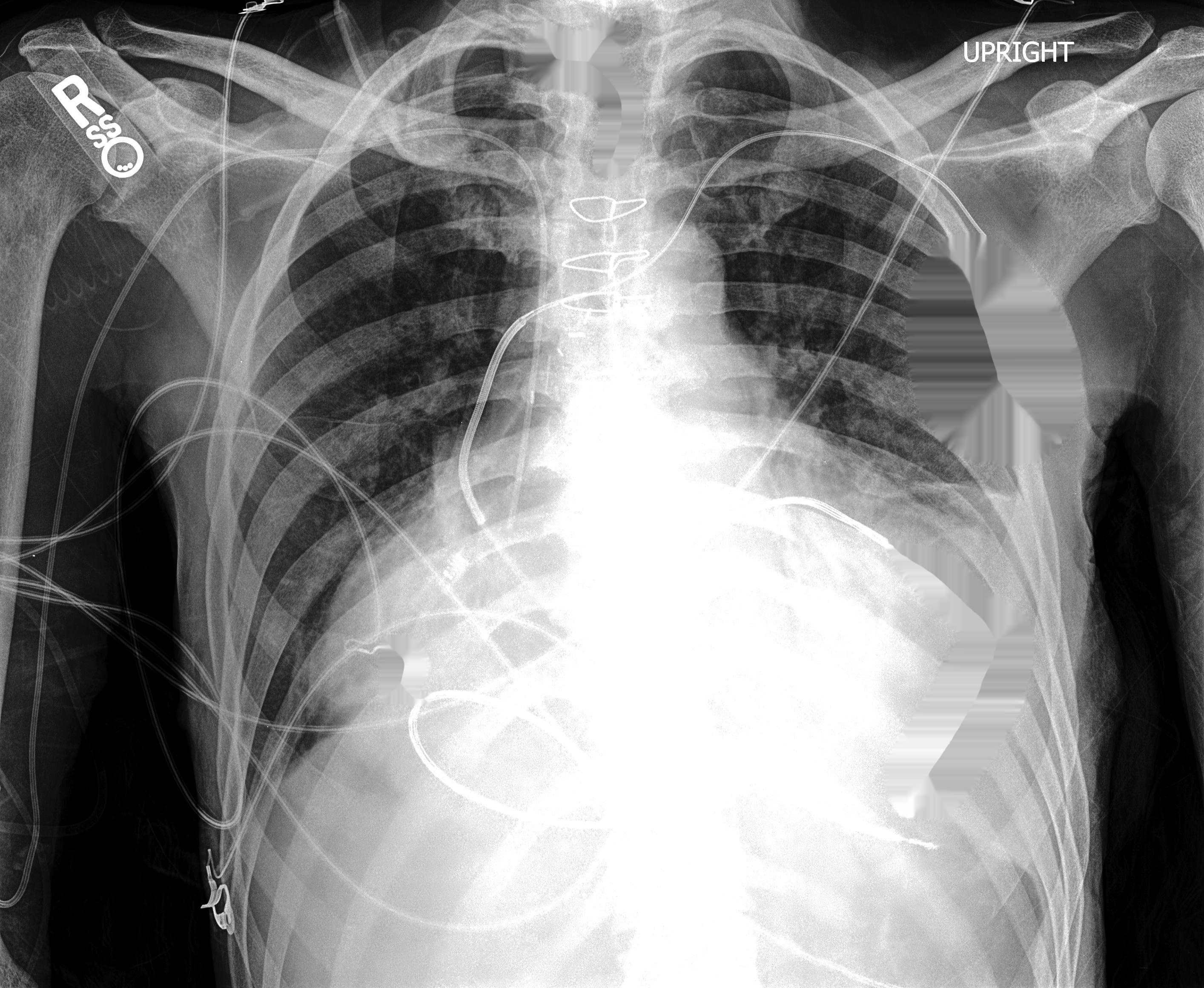}
\end{subfigure}
\begin{subfigure}{.5\textwidth}
  \centering
  \includegraphics[width=.8\linewidth, height=0.7\linewidth]{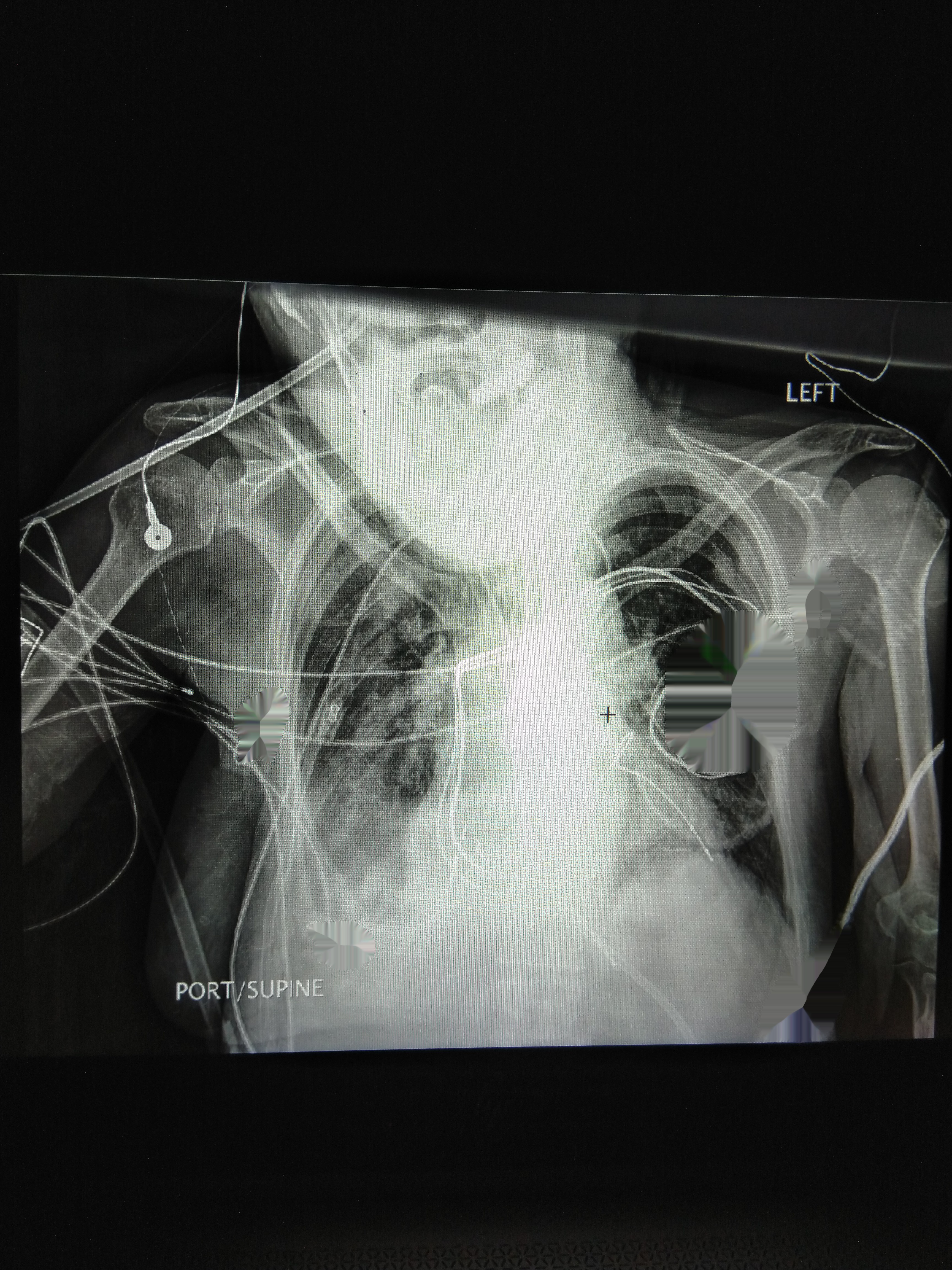}
\end{subfigure}
\caption{Inpainting}
\label{fig:Inpainting}
\end{figure}

\section{Discussion and Future work}
\subsection{Discussion}
To conclude, we mainly focused on removing foreign objects in the center of the lung region of chest radiographs. As a result of object detection, our model has detected different types of foreign objects in chest x-ray images captured by smartphone instead of screening or file images. However, because of the evaluation metrics focused on the lung region, false positives were defined as outside the lung region boundary that led to diminishing accuracy. In addition, the big size objects haven’t been supported. Towards segmentation, our model had a high score in predicted bounding boxes. These outcomes strongly supported removing foreign objects in inpainting section.

As mentioned by \cite{Dang2013visual}, the current image inpainting lacks good quantitative assessment metrics. Thereby, instead of directly evaluating the metrics for the image inpainting task, we assess its contribution to the improvement of the lung abnormality detection models. Reported results in table 7 show that, when applying these techniques with different architectures, we witnessed a significant improvement among conventional classification metrics such as accuracy, F1, Recall and Precision.  In details, for the lightest architecture, VGG16, we obtained a 3.3\% improvement in the accuracy score, while for Resnet50-based architecture and Efficient Net-b2-based architecture, that number is 3.8\% and 0.9\% respectively. It is worth noting that, in all cases, using the inpainting technique led to the improvement of F1-scores, precision and recall. In other words, we observe that no trade off between precision and recall happened. It also indicated that the Type I and Type II error are reduced consecutively.
The experimental achievements will probably apply for the automatic detection of complicated images.

\subsection{Limitation}

Although the paper has achieved high performance by removing the foreign object and proposed methodology, we always acknowledge the limitation of our research. The first limitation is that, since this study is the first one in designing a particular solution to handle the foreign objects in the CheXphoto dataset, it is clear that at the time we were writing, there is no other existing study for us to compare. Secondly, it is worth noting that the object detection and segmentation annotations were conducted under our defined regulations and standards. Thus, these standards may be applicable to the current dataset only.  To address this problem, we need to build a gold standard to encompass foreign objects in general, for example, direction, diagonal length, box area, etc. Third, our approach requires several models working subsequently. Thus, it is not a lightweight solution that can be either applied in Edge devices, or in mobile applications. 

\subsection{Future work}

In future work, this model can probably be used in terms of model detection for multi-label lung disease to the cheXphoto dataset.  Moreover, we continually plan to mitigate foreign objects outside the lung region as well as other objects. 

\section{Acknowledgments}
We would like to give a thank to our friends who collected dataset and labeled segmentation. Ms. Nhung H. Tran (University of Medicine and Pharmacy), Ms.Huong T. T. Nguyen (Hanoi National University of Education).

\bibliographystyle{unsrt}
\bibliography{template}  


\end{document}